\documentclass[letterpaper]{article} 
\usepackage{aaai23}  
\usepackage{xcolor}
\usepackage{times}  
\usepackage{helvet}  
\usepackage{courier}  
\usepackage{balance}
\usepackage{amsmath}
\usepackage{tikz}
\usepackage{fontawesome}
\usetikzlibrary{shapes, arrows.meta, positioning, calc, shadows}
\usepackage{multicol}
\usepackage[hyphens]{url}  
\usepackage{graphicx} 
\usepackage{natbib}  
\usepackage{caption} 
\usepackage{amsmath,amssymb}
\usepackage{algorithm}
\usepackage{algorithmic}

\usepackage{newfloat}
\usepackage{listings}
\DeclareCaptionStyle{ruled}{labelfont=normalfont,labelsep=colon,strut=off} 
\floatstyle{ruled}
\newfloat{listing}{tb}{lst}{}
\floatname{listing}{Listing}
\title{Towards Theoretical Understanding of Data-Driven Policy Refinement}
\author {
    Ali Baheri \textsuperscript{}
}
\affiliations {
    \textsuperscript{} Rochester Institute of Technology\\
    akbeme@rit.edu
}

\usepackage{bibentry}

\begin{document}

\maketitle

\begin{abstract}

This paper presents an approach for data-driven policy refinement in reinforcement learning, specifically designed for safety-critical applications. Our methodology leverages the strengths of data-driven optimization and reinforcement learning to enhance policy safety and optimality through iterative refinement. Our principal contribution lies in the mathematical formulation of this data-driven policy refinement concept. This framework systematically improves reinforcement learning policies by learning from counterexamples identified during data-driven verification. Furthermore, we present a series of theorems elucidating key theoretical properties of our approach, including convergence, robustness bounds, generalization error, and resilience to model mismatch. These results not only validate the effectiveness of our methodology but also contribute to a deeper understanding of its behavior in different environments and scenarios.

\end{abstract}

\vspace{-5 mm}
\section{Introduction}

In the pursuit of creating autonomous systems that not only perform optimally but also operate safely in real-world environments, the field of artificial intelligence and robotics has increasingly turned to reinforcement learning (RL). The inherent ability of RL to learn and optimize behaviors through interaction with its environment makes it an attractive methodology for autonomous systems. However, as we continue to deploy these learning-enabled systems in safety-critical domains such as autonomous driving, healthcare, and aerospace, ensuring their safe operation becomes a paramount concern. The risk of unsafe behavior, particularly in uncertain and dynamically changing environments, poses a significant challenge \cite{dulac2019challenges}.

The need for safety assurance in RL, especially in safety-critical applications, has motivated many advances in the field, yet considerable challenges remain \cite{kober2013reinforcement,kiran2021deep,razzaghi2022survey}. Safe RL focuses on designing learning algorithms that consider safety constraints throughout the learning process, not just as an afterthought. These safety constraints could involve minimizing harm to the environment, adhering to specific operational guidelines, or ensuring minimal deviation from expected behavior. Various strategies for safe RL have been proposed, including methods that incorporate safety constraints into the RL objective function, techniques that allow risk-averse exploration, and approaches that provide safety guarantees by leveraging formal methods \cite{garcia2015comprehensive,baheri2020deep,baheri2022safe,jaimungal2022robust,fulton2018safe,konighofer2020shield}. Within the scope of safe RL, the refinement of policies stands as a crucial strategy, especially in safety-critical applications. Policy refinement is an iterative process that focuses on enhancing an agent’s policy to ensure it exhibits both safe and optimal behavior while conforming to environmental constraints and task specifications. It entails the evaluation of an agent’s present policy, the identification of unsafe or suboptimal actions, and the subsequent update of the policy to mitigate these deficiencies. Thus, policy refinement not only ensures the safety of the RL system during both learning and deployment phases but also continually improves its performance

Two techniques, counterexample-guided abstraction refinement (CEGAR) and counterexample-guided inductive synthesis (CEGIS), have gained traction due to their robust approach to system refinement. CEGAR, originally conceived for the formal verification of finite-state systems, operates by iteratively refining an abstract system model based on counterexamples unearthed during verification \cite{clarke2000counterexample}. While CEGAR hasn't been widely used in RL, its principles of abstract representation and iterative refinement through counterexamples could be insightful for policy analysis and refinement in RL \cite{jin2022trainify}. Similarly, CEGIS is a methodology centered on the synthesis of a correct-by-construction program or policy satisfying a given specification \cite{sketch06}. It starts with an initial candidate policy or program, which is then refined based on counterexamples encountered during the verification phase. CEGIS has shown its effectiveness in various domains such as program synthesis \cite{solar2008program,alur2013syntax} and controller synthesis \cite{henzinger2003counterexample,ravanbakhsh2016robust}. 

Both CEGAR and CEGIS underscore the potential role of counterexamples in refining a policy, opening up new avenues for exploration in RL. By integrating the principles of these techniques with safe RL strategies, there is potential to create a more efficient policy refinement process, contributing to safer and more reliable learning-enabled systems for safety-critical applications. These methodologies underscore the importance of counterexample-based techniques in enhancing the safety and optimality of RL policies and hint at the potential for further advancements in the field of policy refinement. 


This work aims to lay the foundation for \emph{data-driven policy refinement} in reinforcement learning. Our primary goal is to bridge the gap between data-driven verification and reinforcement learning, offering an iterative approach that synergizes the strengths of both domains. In our proposed approach, we propose a data-driven verification method that identifies unsafe trajectories in the current policy. This verification stage acts as a performance check, ensuring that the policy meets the necessary safety and optimality criteria. However, rather than treating the verification stage as an endpoint, we leverage the identified counterexamples as crucial learning opportunities. In essence, the verification stage drives the learning process, providing direct, actionable feedback on the policy's performance. Following the data-driven verification, we use a learning approach that specifically targets the identified counterexamples. This learning stage takes the counterexamples as input, treating them as problem areas that the policy needs to address. Through learning from these counterexamples, the policy is refined iteratively, gradually improving its safety and optimality.

This paper contributes to the field of RL theory, especially in the area of policy refinement. We propose a mathematical formulation of data-driven policy refinement. This approach intertwines the principles of data-driven optimization with reinforcement learning. Through this formulation, we present a methodology for iteratively refining an RL policy. This refinement is guided by counterexamples identified through data-driven verification. Our approach offers a systematic way to enhance the safety and optimality of reinforcement learning policies, providing a robust framework that adapts and improves with each iteration. In addition to introducing the concept of data-driven policy refinement, we delve into the theoretical aspects that underpin this approach. We present a series of theorems that elucidate the implications of data-driven refinement. These theorems enhance our theoretical understanding of data-driven policy refinement. They provide insights into the convergence, generalization, and adaptability of the process, offering theoretical support for our approach.

\section{Problem Formulation and Methodology}

The ultimate objective of this work is to addresses the task of \emph{refining} an optimized policy to ensure it fulfills a predefined safety specification. Our setting comprises a policy parameter space, $\Theta$, which represents all feasible policy parameters that define an agent's behavior in a specific environment. These policies are parameterized by a vector $\theta \in \Theta$. The safety specification, $\varphi$, is a formal prescription of the agent's desired safety behavior. We quantify the satisfaction of the safety specification by a robustness value function $g(\theta; {e})$, which measures the extent to which the safety specification is satisfied for a given policy parameter $\theta$ in a particular environment configuration ${e}$. We also consider the set of possible environment configurations, $\mathcal{E}$, which represent the different conditions the agent may encounter. Our proposed methodology is an iterative process alternating between a data-driven verification approach, using Bayesian optimization, and policy refinement through inverse RL (IRL). The steps are as follows:

\noindent\textbf{Data-Driven Verification via Bayesian Optimization.} Bayesian optimization (BO) is a powerful method for globally optimizing black-box functions that are computationally expensive to evaluate. BO operates by constructing a probabilistic model of the objective function and subsequently using this model to identify promising points for evaluation \cite{snoek2012practical}. Owing to its effectiveness, BO has been employed in a wide array of fields, including hyperparameter tuning in machine learning \cite{wu2019hyperparameter}, design optimization in engineering \cite{garnett2010bayesian}, and decision-making under uncertainty \cite{baheri2017real}.

Our process begins with an initial optimized policy $\theta_1 \in \Theta$. The objective of this phase is to identify environments ${e}^{*}(\theta_i)$ where the policy, characterized by parameters $\theta_i$, infringes the safety specification. To this end, we used BO to uncover the environment configuration that minimizes the robustness value function $g(\theta_i; \mathbf{e})$ for each iteration $i=1,2,\dots$:

\begin{equation}
{e}^{*}(\theta_i) = \underset{\mathbf{e} \in \mathcal{E}}{\mathrm{argmin}} \ g(\theta_i; {e}).
\end{equation}
In each iteration of the BO process, we simulate the agent's behavior in the environment configuration $e$ that minimizes the robustness value function, under the current policy parameters $\theta_i$. The agent's behavior generates a trajectory $\xi$ that represents the sequence of states and actions taken by the agent in the environment configuration ${e}$. If the trajectory $\xi$ violates the safety specification (i.e., $g(\theta_i; {e})<0$), it is considered as a counterexample. The trajectory $\xi$ provides specific instances where the policy with parameters $\theta_i$ fails to meet the safety specification in the corresponding environment configuration ${e}$. This counterexample then serves as the input to the IRL process for policy refinement. In this way, the data-driven verification step leverages the power of BO to systematically explore the environment configuration space and identify counterexamples that challenge the safety of the current policy.

\noindent\textbf{Policy Refinement via Inverse RL.} In this stage, we refine the policy by updating the parameters $\theta_i$ based on the generated counterexamples $e^{*}(\theta_i)$ using IRL. The update yields a new policy parameterized by $\theta_{i+1}$:

\begin{equation}
\theta_{i+1} = \mathrm{IRL}(\theta_i, {e}^{*}(\theta_i)) \quad \text{for } i=1,2,\dots.
\end{equation}
The goal of this IRL process is to refine the policy parameters $\theta$ such that the updated policy $\pi(\theta)$ minimizes the likelihood of producing unsafe trajectories. This is achieved by formulating an optimization problem that minimizes the reward associated with unsafe trajectories and maximizes it for safe ones:

\begin{equation}
\theta_{i+1} = \underset{\theta \in \Theta}{\mathrm{argmin}} \ \mathbb{E}_{\pi(\theta)}[\mathcal{R}(\xi)] - \lambda \mathbb{E}_{\pi(\theta)}[\log(\pi(\theta))],
\end{equation}
where $\mathcal{R}(\xi)$ is the reward function assigning low rewards to unsafe trajectories $\xi$ and high rewards to safe trajectories, $\pi(\theta)$ denotes the policy induced by parameter $\theta$, and $\lambda > 0$ is a trade-off parameter balancing between reward maximization and entropy maximization for adequate exploration. The first term represents the expected reward of trajectories $\xi$ when following the policy $\pi(\theta)$. The reward function $\mathcal{R}(\xi)$ assigns low rewards to unsafe trajectories (counterexamples) and high rewards to safe trajectories. Therefore, by minimizing this term, the policy parameters $\theta$ are updated to discourage unsafe behaviors that lead to low rewards. The second term is used to encourage exploration of different actions. The entropy of a policy is a measure of its randomness, so maximizing the entropy encourages the policy to be more uncertain or random, which in turn encourages exploration of different actions. The parameter $\lambda > 0$ is a trade-off parameter that balances the two objectives: minimizing the expected reward and maximizing the entropy.

The new policy parameters $\theta_{i+1}$ are then used in the next BO iteration to find new counterexamples. The iterative process between BO and IRL continues until a policy parameter $\theta^* \in \Theta$ is found such that $g(\theta^*) \geq 0$, indicating that the safety specification $\varphi$ is satisfied by the policy with parameters $\theta^*$. This iterative process leverages the strengths of both BO and IRL: BO efficiently explores the environment configuration space and identifies counterexamples, while IRL learns a reward function that guides the policy towards avoiding unsafe behaviors.

\begin{figure}[t]
 \centering
  \resizebox{\columnwidth}{!}{
\begin{tikzpicture}[
    block/.style={rectangle, rounded corners, draw, minimum width=4cm, minimum height=2.5cm, align=center, top color=blue!20, bottom color=blue!20, drop shadow, font=\LARGE},
    arrow/.style={-Stealth, line width=5pt, shorten >=2pt, shorten <=2pt},
    label/.style={font=\LARGE, align=center},
    scale=1.5, every node/.style={transform shape}
]
\node[block] (init) {Initial Optimized Policy};
\node[block, right=4cm of init] (bo) {Data-Driven Counterexample \\ Generation (Bayesian Optimization)};
\node[block, right=4cm of bo] (irl) {Policy Refinement\\ (Inverse Reinforcement Learning};

\draw[arrow] (init) -- (bo);
\draw[arrow] (bo) -- (irl) node[midway, above, label] {Counterexamples};
\draw[arrow] (irl) --++ (0,-2.5) -| node[midway, below, label] {Refined Policy} (bo);

\end{tikzpicture}}%
\caption{The iterative process of data-driven verification and inverse RL. The process begins with an optimized policy. A data-driven verification approach is then used for counterexample generation, which serves as input for policy refinement via inverse RL. The refined policy then feeds back into the verification stage, creating a loop for continual policy improvement.}
\label{fig:BO-IRL}
\end{figure}
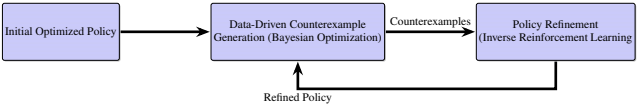

Once the iterative process terminates, it is essential to verify the refined policy $\theta^*$ against the safety specification $\varphi$ using formal verification techniques. This ensures that the refined policy indeed satisfies the safety requirements in various environment configurations, providing a higher level of confidence in the safety of the agent's behavior. The overall methodology is illustrated in Fig. \ref{fig:BO-IRL}. 

\section{Theoretical Results}

In this section, we delve into the theoretical foundations that support our proposed data-driven policy refinement approach. These foundations, laid out in a series of theorems, provide rigorous mathematical evidence that validates the principles and effectiveness of our methodology. Specifically, we explore five key areas:

\begin{enumerate}
    \item The Convergence of Policy Refinement using Counterexamples Theorem, which establishes that our iterative policy refinement process will indeed converge to a policy that satisfies the safety specification.
    \item The Robustness Value Function Bounds Theorem, which provides bounds on the robustness of the value function, thus quantifying the effectiveness of our approach in maintaining safety constraints.
    \item The BO-IRL Generalization Error Bound Theorem, which gives us an upper limit on the discrepancy between the learned policy's performance and the optimal policy's performance, thus indicating how well our approach generalizes.
    \item The Convergence Rate Theorem, which quantifies the rate at which our iterative refinement process converges to a safe policy.
    \item The Robustness of BO-IRL to Model Mismatch Theorem, which shows the resilience of our approach to discrepancies between the estimated and true environment models.
\end{enumerate}
These theorems sheds light on crucial properties of our proposed approach, strengthening the theoretical underpinnings of our work. In the following, we will present a detailed discussion of each theorem, along with their proofs and implications for our data-driven policy refinement approach

\textbf{Theorem 1. (Convergence of Policy Refinement using Counterexamples)}
\textit{Given a set of counterexamples $C$ generated by the BO process for falsifying the learning-based policy, and assuming the IRL algorithm refines the policy using these counterexamples, the refined policy $\pi'(a|s;\theta')$ will converge to a policy that satisfies the safety specification $\varphi$.}

\textbf{PROOF.} We define a robustness value function $g(\theta)$ that measures the robustness of a policy $\pi(a|s;\theta)$ with parameters $\theta$ with respect to the safety specification $\varphi$. For example, $g(\theta)$ could be the minimum robustness value $\rho_{\varphi}$ of the policy $\pi(a|s;\theta)$ over the state-action space. At each iteration $k$, the IRL algorithm uses the counterexamples $C_k$ to compute a better policy $\pi_k(a|s;\theta_k)$. Let $J(\theta_k) = \mathbb{E}[\sum_{t=0}^{T} r(s_t, a_t)|\pi_k(a|s;\theta_k)]$ be the expected cumulative reward of the policy $\pi_k(a|s;\theta_k)$, and let $g_k(\theta_k)$ be the robustness value function for the same policy. The IRL algorithm aims to maximize the expected cumulative reward while ensuring the robustness value function satisfies the safety specification:

\begin{equation}
 \theta_{k+1}=\underset{\theta}{\operatorname{argmax}} J(\theta) \text { subject to } g_k(\theta) \geq \rho_{\varphi}(\theta)   
\end{equation}
As the IRL algorithm refines the policy using the counterexamples, the new policy $\pi_{k+1}(a|s;\theta_{k+1})$ will have a higher robustness value function $g_{k+1}(\theta_{k+1})$ compared to the previous policy $\pi_k(a|s;\theta_k)$:

\begin{equation}
    g_{k+1}\left(\theta_{k+1}\right) \geq g_k\left(\theta_k\right)
    \label{eqn:step2}
\end{equation}
To prove the convergence of the refined policy, we need to show that the sequence of policies generated during each iteration monotonically improves the robustness value function $g(\theta)$. From Eq. \ref{eqn:step2}, we have:

\begin{equation}
g_{k+1}\left(\theta_{k+1}\right) \geq g_k\left(\theta_k\right) \geq g_{k-1}\left(\theta_{k-1}\right) \geq \cdots \geq g_0\left(\theta_0\right)
\end{equation}
Since $g(\theta)$ is non-decreasing with respect to the iteration index $k$, it will have a limit point, say $g^*(\theta^*)$. Using the limit point, we can prove that the robustness value function converges, i.e., $g(\theta)$ approaches a maximum value as the number of iterations increases:

\begin{equation}
\lim _{k \rightarrow \infty} g_k\left(\theta_k\right)=g^*\left(\theta^*\right)    \end{equation}
To establish convergence, we need to prove that the sequence of policies has a converging subsequence, i.e., there exists a subsequence of policies $\{\pi_{k_i}(a|s;\theta_{k_i})\}_{i=1}^\infty$ such that:

\begin{equation}
  \lim _{i \rightarrow \infty} \pi_{k_i}\left(a \mid s ; \theta_{k_i}\right)=\pi^*\left(a \mid s ; \theta^*\right) 
\end{equation}
Under certain conditions, such as compactness of the space of policy parameters $\theta$ and continuity of the robustness value function $g(\theta)$, we can establish that the sequence of policies converges to a policy that satisfies the safety specification $\varphi$. Now we demonstrate that the refinement process results in a policy with a bounded regret as the number of iterations increases. The regret is defined as the difference between the expected cumulative reward of the optimal safe policy and the expected cumulative reward of the policy obtained by the algorithm. Let $J^*(\theta^*)$ be the expected cumulative reward of the optimal safe policy, and let $J(\theta_k)$ be the expected cumulative reward of the policy at iteration $k$. The regret at iteration $k$ can be defined as: 

\begin{equation}
R_k=J^*\left(\theta^*\right)-J\left(\theta_k\right)
\end{equation}
To show that the regret is bounded, we need to demonstrate that there exists an upper bound for $R_k$ as the number of iterations increases. We've previously shown that the robustness value function $g(\theta)$ converges to a maximum value as the number of iterations increases. Since the IRL algorithm is designed to maximize the expected cumulative reward while ensuring the robustness value function satisfies the safety specification, the difference between the expected cumulative rewards of the optimal safe policy and the policy at iteration $k$ should decrease as the robustness value function converges to its maximum value. In other words, the convergence of the robustness value function implies the convergence of the expected cumulative rewards: $\lim _{k \rightarrow \infty} J\left(\theta_k\right)=J^*\left(\theta^*\right)$. This suggests that the regret $R_k$ approaches zero as the number of iterations increases:

\begin{equation}
 \lim _{k \rightarrow \infty} R_k=J^*\left(\theta^*\right)-J\left(\theta_k\right)=0  
\end{equation}
Since the regret $R_k$ converges to zero, it implies that there exists an upper bound for the regret, and the regret is indeed bounded as the number of iterations increases. This shows that the refined policy obtained by the proposed refinement approach converges to a policy that not only satisfies the safety specification but also has a bounded regret compared to the optimal safe policy.

\textbf{Theorem 2. (Bounds on Robustness).} \textit{If the robustness value function $g(\theta; e)$ is Lipschitz continuous with Lipschitz constant $L_g$, and there exists a constant $C_{\text{IRL}} > 0$ such that $|\theta_{i+1} - \theta_i| \leq C_{\text{IRL}}$ for all $i = 1,2,\dots$, then after $n$ iterations of the process, the lower bound on the robustness value function for the final policy $\theta^*$ is given by:}

\begin{equation}
  g\left(\theta^* ; e\right) \geq g\left(\theta_1 ; e\right)-L_g C_{\mathrm{IRL}}(n-1)  
\end{equation}

\textbf{PROOF.} By the Lipschitz continuity of the robustness value function $g(\theta; e)$, we know that for any two policy parameters $\theta, \theta' \in \Theta$ and environment configuration $e \in \mathcal{E}$, the following inequality holds:

\begin{equation}
    \left|g(\theta ; e)-g\left(\theta^{\prime} ; e\right)\right| \leq L_g \| \theta-\theta^{\prime} \mid
    \label{T1_lipschitz}
\end{equation}
In our iterative process, we are given that $|\theta_{i+1} - \theta_i| \leq C_{\text{IRL}}$ for all $i = 1,2,\dots$. Substituting this into the inequality \ref{T1_lipschitz}, we get:

\begin{equation}
    \left|g\left(\theta_{i+1} ; e\right)-g\left(\theta_i ; e\right)\right| \leq L_g\left\|\theta_{i+1}-\theta_i\right\| \leq L_g C_{\mathrm{IRL}}
\end{equation}
for all $i = 1,2,\dots$. Since the policy is refined iteratively using IRL, we have $g(\theta_{i+1}; e) \geq g(\theta_i; e)$ for all $i = 1,2,\dots$. Therefore,

\begin{equation}
    g\left(\theta_{i+1} ; e\right)-g\left(\theta_i ; e\right) \geq-L_g C_{\mathrm{IRL}}
    \label{T1_step3}
\end{equation}
Now, we sum the inequalities from \ref{T1_step3} for each iteration from $i = 1$ to $i = n-1$

\begin{equation}
  \sum_{i=1}^{n-1}\left(g\left(\theta_{i+1} ; e\right)-g\left(\theta_i ; e\right)\right) \geq-L_g C_{\mathrm{IRL}}(n-1)  
\end{equation}
One can observe that the summation on the left-hand side forms a telescoping series, which simplifies as

\begin{equation}
    g\left(\theta^* ; e\right)-g\left(\theta_1 ; e\right) \geq-L_g C_{\mathrm{IRL}}(n-1)
\end{equation}
Rearranging the inequality, we obtain the desired lower bound on the robustness value function for the final policy $\theta^*$:

\begin{equation}
    g\left(\theta^* ; e\right) \geq g\left(\theta_1 ; e\right)-L_g C_{\mathrm{IRL}}(n-1)
\end{equation}
This completes the proof. $\blacksquare$

The theorem establishes a lower bound on the robustness value function for the final refined policy $\theta^*$ after $n$ iterations of the process. The bound depends on the initial policy's robustness value, the Lipschitz constant $L_g$ of the robustness value function, and the constant $C_{\text{IRL}}$ that captures the maximum change in the policy parameters during each iteration. This result provides insights into how close the refined policy is to fully satisfying the safety specification and whether it can be further improved.

\noindent{\textbf{Theorem 3. (BO-IRL Generalization Error Bound).} \textit{Given a dataset $\mathcal{D}$ consisting of $N$ i.i.d. samples, a confidence level $1 - \delta$, and a BO-IRL algorithm with $K$ iterations, the generalization error $\epsilon$ of the policy obtained by the BO-IRL algorithm is bounded as follows with probability at least $1 - \delta$:}

$$\epsilon \leq 2 K \operatorname{Rad}(\mathcal{D})+\sqrt{\frac{8 K \log (1 / \delta)}{N}}$$
\textit{where $\operatorname{Rad}(\mathcal{D})$ is the Rademacher complexity of the dataset.}

\textbf{PROOF.} The generalization error $\epsilon$ is the difference between the expected cumulative reward of the learned policy on the dataset $\mathcal{D}$ and the expected cumulative reward of the same policy on the true distribution of the environment:

\begin{equation}
  \epsilon=\left|\mathbb{E}_{\mathcal{D}}[J(\theta)]-\mathbb{E}_{\text {true }}[J(\theta)]\right| 
\end{equation}
Now we define the empirical Rademacher averages for the dataset $\mathcal{D}$ as:

\begin{equation}
  \hat{\mathcal{R}}_N(\mathcal{D})=\mathbb{E}_\sigma\left[\sup _\theta \frac{1}{N} \sum_{i=1}^N \sigma_i J\left(\theta_i\right)\right] 
\end{equation}
where $\sigma_i \in {-1, 1}$ are independent Rademacher random variables, and $\theta_i$ are the policy parameters associated with the $i$-th sample in the dataset. The Rademacher complexity $\operatorname{Rad}(\mathcal{D})$ can be computed as the expected value of the empirical Rademacher averages: $\operatorname{Rad}(\mathcal{D}) = \mathbb{E}_{\mathcal{D}}\left[\hat{\mathcal{R}}_N(\mathcal{D})\right]$. For each iteration of the BO-IRL algorithm, we can apply the Rademacher generalization bound. For any $\delta > 0$, with probability at least $1 - \frac{\delta}{K}$:

\begin{equation}
\epsilon_k \leq 2 \operatorname{Rad}(\mathcal{D})+\sqrt{\frac{2 \log (K / \delta)}{N}}
\end{equation}
where $\epsilon_k$ is the generalization error for the policy at iteration $k$. The total generalization error after $K$ iterations can be obtained by summing the individual generalization errors and applying the union bound

\begin{equation}
\epsilon \leq \sum_{k=1}^K \epsilon_k \leq 2 K \operatorname{Rad}(\mathcal{D})+\sqrt{\frac{8 K \log (1 / \delta)}{N}}
\end{equation}
with probability at least $1 - \delta$. This concludes the proof of the BO-IRL generalization error bound. $\blacksquare$ 

The theorem provides a bound on the performance of the refined policy on unseen situations, taking into account the iterative nature of the BO-IRL algorithm. The generalization error bound helps us understand the relationship between the number of iterations, the amount of data, and the generalization performance of the learned policy.

\textbf{Theorem 4. (Convergence Rate).} \textit{Under certain assumptions about the problem setup, the exploration-exploitation trade-off in the BO process, the quality of the counterexamples, and the policy improvement in the IRL, the combined BO-IRL approach converges to a safe policy that satisfies the given safety specification at a rate determined by the relationship between the policy improvement metric $\Delta_k$ and the number of IRL iterations $k$.}

\textbf{PROOF [Sketch].} We assume that the problem setup, the exploration-exploitation trade-off in the BO process, and the policy improvement in the IRL are such that the policy improvement metric $\Delta_k$ decreases with the number of IRL iterations $k$. This relationship can be represented as: $\Delta_k=f(k)$ where $f(k)$ is a monotonically decreasing function of $k$. Due to the combined BO-IRL approach, the policy is iteratively improved using counterexamples generated by the BO process and refined by the IRL. The policy improvement can be quantified by the policy improvement metric $\Delta_k$, which measures the difference between the expected cumulative rewards of two consecutive policies:

\begin{equation}
  \Delta_k=J\left(\theta_{k+1}\right)-J\left(\theta_k\right)  
\end{equation}
where $J(\theta_k)$ and $J(\theta_{k+1})$ represent the expected cumulative rewards of the policies at iterations $k$ and $k+1$, respectively. Since the policy improvement metric $\Delta_k$ decreases with the number of IRL iterations $k$, it implies that the policy is improving at each iteration. As the number of iterations increases, the policy improvement metric $\Delta_k$ will approach zero, indicating that the expected cumulative rewards are no longer significantly improving:

\begin{equation}
\lim _{k \rightarrow \infty} \Delta_k=0
\end{equation}
The rate of convergence is determined by the relationship between the policy improvement metric $\Delta_k$ and the number of IRL iterations $k$. In other words, the speed at which the combined BO-IRL approach converges to a safe policy that satisfies the given safety specification is governed by how fast the function $f(k)$ decreases with increasing $k$. This proof sketch shows that the convergence rate of the combined BO-IRL approach is determined by the relationship between the policy improvement metric $\Delta_k$ and the number of IRL iterations $k$. 

\textbf{Theorem 5. (Robustness of BO-IRL to Model Mismatch).} \textit{Given any $\epsilon > 0$, if $|\mathcal{M}^* - \mathcal{M}| \leq \epsilon$, then for any policy $\pi_\theta$, the difference in the expected reward under the true model $\mathcal{M}^*$ and the estimated model $\mathcal{M}$, denoted as $\Delta J(\pi_\theta)$, is bounded by $\epsilon$.}

\textbf{PROOF.} Let $\mathcal{M}^*$ denote the true environment model and $\mathcal{M}$ the model used by the BO-IRL algorithm. We can define the model-induced policy performance mismatch as:

\begin{equation}
\Delta J(\pi_\theta) = |J(\pi_\theta; \mathcal{M}^*) - J(\pi_\theta; \mathcal{M})|,
\end{equation}
where $J(\pi_\theta; \mathcal{M})$ is the expected cumulative reward of policy $\pi_\theta$ under model $\mathcal{M}$. Our goal is to show that $\Delta J(\pi_\theta)$ is bounded by $\epsilon$. We can express $\Delta J(\pi_\theta)$ as:

\begin{equation}
\Delta J(\pi_\theta) = \left|\mathbb{E}_{\xi \sim \pi_\theta, \mathcal{M}^*}[\mathcal{R}(\xi)] - \mathbb{E}_{\xi \sim \pi_\theta, \mathcal{M}}[\mathcal{R}(\xi)]\right|,
\end{equation}
where $\mathcal{R}(\xi)$ represents the reward function. Now, we use the assumption that $|\mathcal{M}^* - \mathcal{M}| \leq \epsilon$. Given this assumption, the difference between the expected rewards under the true model and the estimated model is also bounded by $\epsilon$. Hence, we have

\begin{equation}
\Delta J(\pi_\theta) = \left|\mathbb{E}_{\xi \sim \pi_\theta, \mathcal{M}^*}[\mathcal{R}(\xi)] - \mathbb{E}_{\xi \sim \pi_\theta, \mathcal{M}}[\mathcal{R}(\xi)]\right| \leq \epsilon
\end{equation}
Therefore, the difference in the expected reward under the true model $\mathcal{M}^*$ and the estimated model $\mathcal{M}$ for any policy $\pi_\theta$ is bounded by $\epsilon$. $\blacksquare$

The theorem provides a guideline for the refinement process by bounding the difference in expected rewards between the true and estimated models as $\epsilon$. This ensures that even if a policy performs poorly in the estimated model, its performance will improve in the true model through refinement, as long as the model mismatch stays within the $\epsilon$ limit. As a result, the refined policies become robust, as the theorem guarantees that their deviation from expected outcomes in the true model will not exceed $\epsilon$. This acts as a safeguard against overfitting to the estimated model during the refinement stage.


\section{Conclusions and Future Directions}

In this work, we propose an approach for policy refinement in reinforcement learning, particularly for safety-critical applications. By uniquely blending Bayesian optimization (BO) and inverse reinforcement learning (IRL), we have developed a methodology that iteratively refines policies using counterexamples derived from data-driven verification. Furthermore, we present a series of theorems that provide a deeper understanding of the data-driven policy refinement process, revealing key insights into its convergence, bounds on robustness, generalizability, and convergence rate. This theoretical foundation serves as a solid basis for our approach and contributes to the broader RL theory. Future work could explore additional theoretical properties and investigate more efficient algorithms for counterexample generation and policy refinement. Moreover, applying our methodology to real-world problems and evaluating its performance in practice are important avenues for future research.

\vspace{-5 mm}
\bibliography{aaai23}

\end{document}